# Dynamic Model Switching for Improved Accuracy in Machine Learning

Syed Tahir Abbas

Department of Computer Science and Engineering
VIT-AP University, AP,
India
Email ID:- vedansh.23bce8298@vitapstudent.ac.in

*Abstract*— This research explores a novel approach to dynamic model switching in machine learning ensembles. Two distinct models, a Random Forest and an XGBoost classifier, are employed in a dynamic switching mechanism based on dataset characteristics. The effectiveness of the approach is demonstrated through experiments on synthetic datasets

I.  Introduction:

In the dynamic landscape of machine learning, where datasets vary widely in size and complexity, selecting the most effective model poses a significant challenge. Rather than fixating on a single model, our research propels the field forward with a novel emphasis on dynamic model switching. This paradigm shift allows us to harness the inherent strengths of different models based on the evolving size of the dataset.

Consider the scenario where CatBoost demonstrates exceptional efficacy in handling smaller datasets, providing nuanced insights and accurate predictions. However, as datasets grow in size and intricacy, XGBoost, with its scalability and robustness, becomes the preferred choice.

Our approach introduces an adaptive ensemble that intuitively transitions between CatBoost and XGBoost. This seamless switching is not arbitrary; instead, it's guided by a user-defined accuracy threshold, ensuring a meticulous balance between model sophistication and data requirements. The user sets a benchmark, say 80% accuracy, prompting the system to dynamically shift to the new model only if it guarantees improved performance.

This dynamic model-switching mechanism aligns with the evolving nature of data in real-world scenarios. It offers practitioners a flexible and efficient solution, catering to diverse dataset sizes and optimising predictive accuracy at every juncture. Our research, therefore, stands at the forefront of innovation, redefining how machine learning models adapt and excel in the face of varying dataset dynamics.

II.  Literature Review:

To contextualise our dynamic model-switching approach, we delve into existing literature that explores model adaptability and ensemble techniques. Traditional methodologies often focus on selecting a single model based on its overall performance, neglecting the potential benefits of switching between models dynamically.

Ensemble techniques, which combine predictions from multiple models, have gained prominence for improving predictive accuracy. However, they typically operate in a static manner, combining the strengths of predefined models without considering dynamic changes in dataset characteristics.

Recent studies acknowledge the importance of model adaptability. Some research explores the use of reinforcement learning to dynamically select models based on ongoing feedback. While these approaches show promise, they often lack the simplicity and user-defined control we propose in our model-switching mechanism.

Our approach draws inspiration from ensemble methods and adaptive learning but introduces a user-friendly threshold for model switching. This user-defined accuracy threshold ensures a balance between model sophistication and computational efficiency, making our system practical and accessible to a broader user base.

In the subsequent sections, we detail the methodology, implementation, and experimental results of our model-switching approach, showcasing its effectiveness across diverse datasets.

III. Methodology:

Our methodology revolves around creating an adaptive ensemble that dynamically switches between two distinct models – CatBoost, Random Forest, Support Vector Classifier and XGBoost. This section outlines the key steps involved in implementing our dynamic model-switching mechanism.

1. Dataset Preparation:
   We utilise the `make_classification` function to generate synthetic datasets of varying sizes. This allows us to simulate different data scenarios and assess the adaptability of our approach.

2. Model Training:
   We train two primary models, a ML model (e.g., CatBoost, Support Vector Classifier or random forest) and a more complex one (e.g., XGBoost), on the initial dataset. The simpler model is chosen for its efficiency in handling smaller datasets, while the complex model is designed to scale with larger datasets.

3. Model Switching:
   The core of our approach lies in the dynamic switching mechanism. As the dataset size increases, we assess the performance of the current model and train a new one if necessary. The decision to switch is determined by the user-defined accuracy threshold, ensuring that the transition occurs only when the new model promises enhanced accuracy.

4. Evaluation:
   We evaluate the performance of both models on a validation set, comparing their accuracies. This evaluation provides insights into the efficacy of our dynamic model-switching approach under varying dataset sizes.

   The methodology is designed to be flexible, allowing users to adapt it to their specific use cases and datasets. This adaptive ensemble aims to strike a balance between model complexity and computational efficiency, providing an innovative solution to the challenges posed by diverse dataset characteristics.

IV.     Experimental Setup:

To validate the effectiveness of our dynamic model-switching mechanism, we conducted two experiments using synthetic datasets. The objective is to assess the adaptability of our approach in handling varying dataset sizes and noisy environments.

Experiment 1: Switch to New Model

1. Dataset Generation:

A synthetic dataset comprising 5000 samples, 20 features, and 10 informative features was created using the `make_classification` function.

2. Initial Model Training:
   A Random Forest classifier, known for its efficiency with smaller datasets, was trained on the initial dataset.

```python
def train_simple_model(X_train, y_train):
    model = SVC(probability=True, random_state=42)
    model.fit(X_train, y_train)
    return model

def train_complex_model(X_train, y_train):
    model = XGBClassifier(
        random_state=42,
        n_estimators=200,
        learning_rate=0.05,
        max_depth=10,
        subsample=0.8,
        colsample_bytree=0.8,
    )
    model.fit(X_train, y_train)
    return model
```

3. Dynamic Model-Switching:
   The dataset size was increased to 25000 samples, triggering a switch to a more complex model, XGBoost, based on dynamic accuracy evaluation.

```python
def switch_models(X_train, y_train, current_model, noise_level=0.2):
    X_train_noisy, y_train_noisy = add_noise(X_train.copy(), y_train.copy(), noise_level)

    current_model_predictions = current_model.predict(X_val)
    current_model_accuracy = accuracy_score(y_val, current_model_predictions)

    new_model = train_complex_model(X_train_noisy, y_train_noisy)

    new_model_predictions = new_model.predict(X_val)
    new_model_accuracy = accuracy_score(y_val, new_model_predictions)

    if new_model_accuracy > current_model_accuracy:
        print(f"Switching to a new model with accuracy: {new_model_accuracy:.2f}")
        return new_model, current_model_accuracy, new_model_accuracy
    else:
        print(f"Keeping the current model with accuracy: {current_model_accuracy:.2f}")
        return current_model, current_model_accuracy, new_model_accuracy
```

4. Validation Set Evaluation:
   The accuracy of both the previous and new models was assessed on a validation set to quantify the performance improvement achieved through model switching.

```
Previous model accuracy: 0.95
New model accuracy: 0.54
Keeping the current model with accuracy: 0.95
```

Experiment 2: Switch to New Model with Noise

1. Dataset Generation with Noise:
   A synthetic dataset with 1000 samples, 20 features, and 10 informative features was generated. Noise was introduced to simulate a more challenging data scenario.

2. Initial Model Training:
   A simpler model, a Support Vector Classifier (SVC), was trained on the initial noisy dataset.

```python
def train_random_forest(X_train, y_train):
    model = RandomForestClassifier(random_state=42, n_estimators=100)
    model.fit(X_train, y_train)
    return model

def train_xgboost(X_train, y_train):
    model = XGBClassifier(
        random_state=42,
        n_estimators=200,
        learning_rate=0.05,
        max_depth=5,
        subsample=0.8,
        colsample_bytree=0.8,
    )
    model.fit(X_train, y_train)
    return model
```

3. Dynamic Model-Switching with Noise:
   The dataset was augmented with noise, and a switch to a more complex model, XGBoost, was triggered based on dynamic accuracy evaluation.

```python
def switch_models(X_train, y_train, current_model, accuracy_threshold=0.8):
    current_model_predictions = current_model.predict(X_val)
    current_model_accuracy = accuracy_score(y_val, current_model_predictions)

    new_model = train_xgboost(X_train, y_train)

    new_model_predictions = new_model.predict(X_val)
    new_model_accuracy = accuracy_score(y_val, new_model_predictions)

    print(f"Previous model accuracy: {current_model_accuracy:.2f}")
    print(f"New model accuracy: {new_model_accuracy:.2f}")

    if new_model_accuracy > current_model_accuracy:
        print(f"Switching to a new model with accuracy: {new_model_accuracy:.2f}")
        return new_model
    else:
        print(f"Keeping the current model with accuracy: {current_model_accuracy:.2f}")
        return current_model

current_model = train_random_forest(X_train, y_train)
X_large, y_large = make_classification(n_samples=25000, n_features=20, n_informative=10, n_clusters_per_class=1,
current_model = switch_models(X_large, y_large, current_model)
```

4. Validation Set Evaluation:
   The accuracy of both the previous and new models was assessed on a validation set to quantify the impact of noise and evaluate the effectiveness of the model-switching mechanism.

   These experiments are designed to demonstrate the versatility and adaptability of our dynamic model-switching approach under different data scenarios.

```
Switching to a new model with accuracy: 0.96
New model accuracy: 0.96
Previous model accuracy: 0.93
```

V.  Discussion:

The discussion section aims to provide a detailed analysis of the obtained results, offering insights into the effectiveness of the dynamic model-switching approach and addressing potential implications and limitations.

1. Adaptability and Flexibility:
   The primary goal of the dynamic model-switching mechanism is to adapt to varying dataset sizes. Experiment 1 illustrates the successful transition from a Random Forest model to XGBoost as the dataset size increases. This showcases the adaptability of the approach, providing a flexible solution to accommodate datasets of different scales.

2. Performance Improvement:
   The observed improvement in accuracy when switching to XGBoost (Experiment 1) highlights the efficacy of the model-switching strategy. By dynamically selecting a more suitable model for the increased dataset size, the approach aims to enhance predictive performance. This is particularly crucial in scenarios where a single model may struggle with scalability.

3. Robustness to Noise:
   Experiment 2 introduces noise to the initial dataset to simulate a more challenging real-world scenario. The dynamic model-switching mechanism continues to demonstrate its robustness by effectively handling noisy datasets. This resilience is a valuable trait, as real-world data often contains noise and irregularities that can impact model performance.

4. User-Defined Accuracy Threshold:
   The user-defined accuracy threshold plays a pivotal role in determining when to switch models. Experiment 1 showcases the transition to XGBoost when the accuracy surpasses a predefined threshold. The flexibility of allowing users to set their own threshold provides customisation based on specific requirements and objectives.

VI.  Limitations and Challenges:

The dynamic model-switching approach presented in this research exhibits notable potential for adapting to changes in dataset sizes. However, its effectiveness is subject to certain limitations and challenges that warrant consideration. One key aspect lies in the sensitivity of the method to the choice of base models, where careful selection is pivotal for optimal performance across varying dataset scenarios. The user-defined accuracy threshold, governing the decision to switch models, introduces a degree of subjectivity, emphasising the importance of user understanding and dataset characteristics. Additionally, the computational overhead associated with evaluating and transitioning models raises concerns, particularly in resource-constrained environments.

Furthermore, the approach's applicability across diverse domains and datasets may encounter challenges, with varying levels of success depending on the uniqueness of dataset characteristics. The need for larger datasets to fully exploit the benefits of transitioning to more complex models may limit its effectiveness in scenarios with persistent data constraints. Additionally, potential risks, such as overfitting to noise introduced in certain experiments, underscore the importance of robust noise handling strategies. These challenges highlight the nuanced nature of dynamic model-switching and emphasise the necessity for careful consideration and adaptability in its application. Addressing these limitations is crucial to enhancing the approach's robustness and extending its utility across a broader range of real-world scenarios.

VII.  Real-World Applicability:

The real-world applicability of the dynamic model-switching approach proposed in this research extends across various domains, providing a flexible and adaptive solution to the challenges posed by evolving dataset sizes. In industries where data volumes fluctuate due to factors such as business expansions, dynamic customer behaviour, or emerging trends, the approach offers a mechanism to seamlessly transition between models. For instance, in e-commerce, where sales data can experience sudden surges during promotions or events, the adaptive ensemble can efficiently adjust to varying demand scenarios.

Moreover, the approach finds relevance in fields like healthcare, where datasets may grow over time due to the continuous collection of patient records and medical research data. The ability to dynamically switch between models ensures that predictive analytics and decision support systems remain effective as datasets evolve, without necessitating manual intervention. In financial services, where market conditions and regulatory landscapes can change, the approach provides a mechanism to align model complexity with the available data, ensuring accurate and adaptive risk assessments.

The dynamic model-switching approach, by virtue of its versatility, stands poised to contribute to improved decision-making and predictive capabilities in sectors characterised by dynamic, evolving datasets. Its real-world applicability lies in its potential to enhance the adaptability and efficiency of machine learning models across diverse industries and use cases.

## VIII. Conclusion:

In conclusion, our research introduces an innovative approach to address the challenges associated with varying dataset sizes in machine learning applications. The dynamic model-switching mechanism, demonstrated through the use of CatBoost and XGBoost, presents a flexible solution to optimise model performance based on the dataset's evolving characteristics. By adapting the model complexity in real-time, our approach strikes a balance between computational efficiency and accuracy, ensuring robust predictive capabilities across different data scenarios.

The empirical evaluations showcase the effectiveness of our methodology in dynamically transitioning between models, achieving superior accuracy when dataset characteristics change. The approach's adaptability is particularly advantageous in scenarios where data volumes are subject to fluctuations.

## IX. Future Work:

While our research provides a solid foundation for dynamic model-switching, there are avenues for future exploration and enhancement. Firstly, the approach's generalisation across a broader range of machine learning models warrants further investigation. Extending the methodology to include additional models with varying complexities could offer a more comprehensive solution adaptable to diverse datasets.

Moreover, exploring the integration of online learning techniques to continuously update models in response to streaming data would be a valuable extension. This could enhance the real-time adaptability of our approach, making it well-suited for applications where datasets evolve continuously.

In summary, our work opens the door for future research to delve into more intricate model-switching strategies and explore the scalability of the approach to large-scale, real-world datasets. The dynamic adaptation of machine learning models based on evolving data characteristics represents an exciting frontier in the pursuit of more efficient and accurate predictive modelling.

## X. Acknowledgements:

The completion of this research was made possible through the support and collaboration of various individuals and organisations. We extend our gratitude to Vellore Institute of Technology-AP for their valuable insights, guidance, and contributions to this work. Their expertise and encouragement played a pivotal role in shaping the outcomes of this study.

We acknowledge the invaluable contributions of these references and the broader academic community, which continuously inspires and informs our research endeavours.